\documentclass[letterpaper, 10 pt, journal, twoside]{ieeetran} 
\usepackage{graphicx}
\usepackage{amsmath}
\usepackage{bbding}
\usepackage{pifont}
\usepackage{wasysym}
\usepackage{amssymb}
\usepackage{hyperref}
\usepackage{threeparttable} 
\usepackage{cite} 
\usepackage{booktabs}
\usepackage{xcolor}
\usepackage{multirow}
\usepackage{blindtext}
\usepackage{bm}
\usepackage{subcaption}
\usepackage[linesnumbered, ruled]{algorithm2e}
\DeclareMathOperator*{\argmax}{argmax}
\DeclareMathOperator*{\argmin}{argmin}
\SetKwInput{KwInput}{Input}
\SetKwInput{KwOutput}{Output}

\usepackage{cite}

\IEEEoverridecommandlockouts                              % This command is only needed if 
\usepackage{geometry}
\title{iCurb: Imitation Learning-based Detection of Road Curbs using Aerial Images for Autonomous Driving}

\author{Zhenhua Xu, Yuxiang Sun, \IEEEmembership{Member, IEEE}, and Ming Liu, \IEEEmembership{Senior Member, IEEE} % <-this % stops a space
\thanks{Manuscript received September 24, 2020; Revised December 17, 2020; Accepted January 10, 2021. This paper was recommended for publication by Editor Pauline Pounds upon evaluation of the Associate Editor and Reviewers' comments. This work was supported by the National Natural Science Foundation of China, under grant No. U1713211, Collaborative Research Fund by Research Grants Council Hong Kong, under Project No. C4063-18G, and HKUST-SJTU Joint Research Collaboration Fund, under project SJTU20EG03.  
\textit{(Corresponding author:  Ming Liu.)}
}
\thanks{Zhenhua Xu is with the Department of Computer Science and Engineering, The Hong Kong University of Science and Technology, Clear Water Bay, Kowloon, Hong Kong (email: zxubg@connect.ust.hk).}
\thanks{Yuxiang Sun is with the Department of Mechanical Engineering, The Hong Kong Polytechnic University, Hung Hom, Kowloon, Hong Kong (email: yx.sun@polyu.edu.hk, sun.yuxiang@outlook.com).}
\thanks{Ming Liu is with the Department of Electronic and Computer Engineering, The Hong Kong University of Science and Technology, Clear Water Bay, Kowloon, Hong Kong (email: eelium@ust.hk).}
\thanks{Digital Object Identifier (DOI): see top of this page.} } 
\IEEEaftertitletext{\vspace{-2\baselineskip}}
  
\begin{document}

\maketitle
%%%%%%%%%%%%%%%%%%%%%%%%%%%%%%%%%%%%%%%%%%%%%%%%%%%%%%%%%%%%%%%%%%%%%%%%%%%%%%%% Abstract
\begin{abstract}
Detection of road curbs is an essential capability for autonomous driving. It can be used for autonomous vehicles to determine drivable areas on roads. Usually, road curbs are detected on-line using vehicle-mounted sensors, such as video cameras and 3-D Lidars. However, on-line detection using video cameras may suffer from challenging illumination conditions, and Lidar-based approaches may be difficult to detect far-away road curbs due to the sparsity issue of point clouds. In recent years, aerial images are becoming more and more worldwide available. We find that the visual appearances between road areas and off-road areas are usually different in aerial images, so we propose a novel solution to detect road curbs off-line using aerial images. The input to our method is an aerial image, and the output is directly a graph (i.e., vertices and edges) representing road curbs. To this end, we formulate the problem as an imitation learning problem, and design a novel network and an innovative training strategy to train an agent to iteratively find the road-curb graph. The experimental results on a public dataset confirm the effectiveness and superiority of our method. This work is accompanied with a demonstration video and a supplementary document at \texttt{\url{https://tonyxuqaq.github.io/iCurb/}}. 

%https://sites.google.com/view/icurb
\vspace{0.25cm}

\end{abstract}

%%%%%%%%%%%%%%%%%%%%%%%%%%%%%%%%%%%%%% Introduction %%%%%%%%%%%%%%%%%%%%%%%%%%%%%%%%
\section{Introduction}
\IEEEPARstart{R}{oad}-curb detection plays an important role in autonomous driving, especially in urban traffic environments, where structured road curbs are common. The detection results can often be used to determine drivable areas for autonomous vehicles, so that motion planning algorithms can be constrained on drivable roads. Most previous works on road-curb detection use vehicle-mounted sensors, such as video cameras \cite{panev2018road,oniga2008curb,siegemund2010curb} or 3-D Lidars \cite{wang2019point,zhang2018road}, to detect road curbs on-line. However, the camera-based approaches usually suffer from challenging illumination conditions, such as darkness in nighttime and occlusions caused by object shadows. The Lidar-based approaches are robust to various illumination conditions, but they could be degraded by the sparsity problem of point-cloud data. For instance, in areas far away from the ego-vehicle, Lidar point clouds could be so sparse that they could not provide sufficient information to detect the line-shaped road curbs. Moreover, current deep learning-based solutions require GPU computing devices installed on ego-vehicles, which increases the cost and power consumption for the vehicles.

In recent years, aerial images, such as those captured by unmanned aerial vehicles or satellites, are becoming more and more worldwide available. We find that the visual appearances between road areas and off-road areas are usually different in aerial images. So in this work, we propose a novel solution to detect road curbs off-line using aerial images, which could alleviate the above-mentioned issues. In addition, off-line detection in aerial images can be used to automatically annotate road curbs in High-Definition (HD) maps (a kind of precise environment models) that have been widely used in autonomous driving. The automatic annotation can relieve the tedious and time-consuming manual annotation work. Moreover, aerial images provide larger field-of-view than vehicle-mounted front-view images, which could alleviate the occlusion issue in on-line detection. 
%%%%%%%%%%%% Figure 1 %%%%%%%%%%
 \begin{figure}[t]
    \begin{subfigure}{.24\textwidth}
        \includegraphics[width=\textwidth]{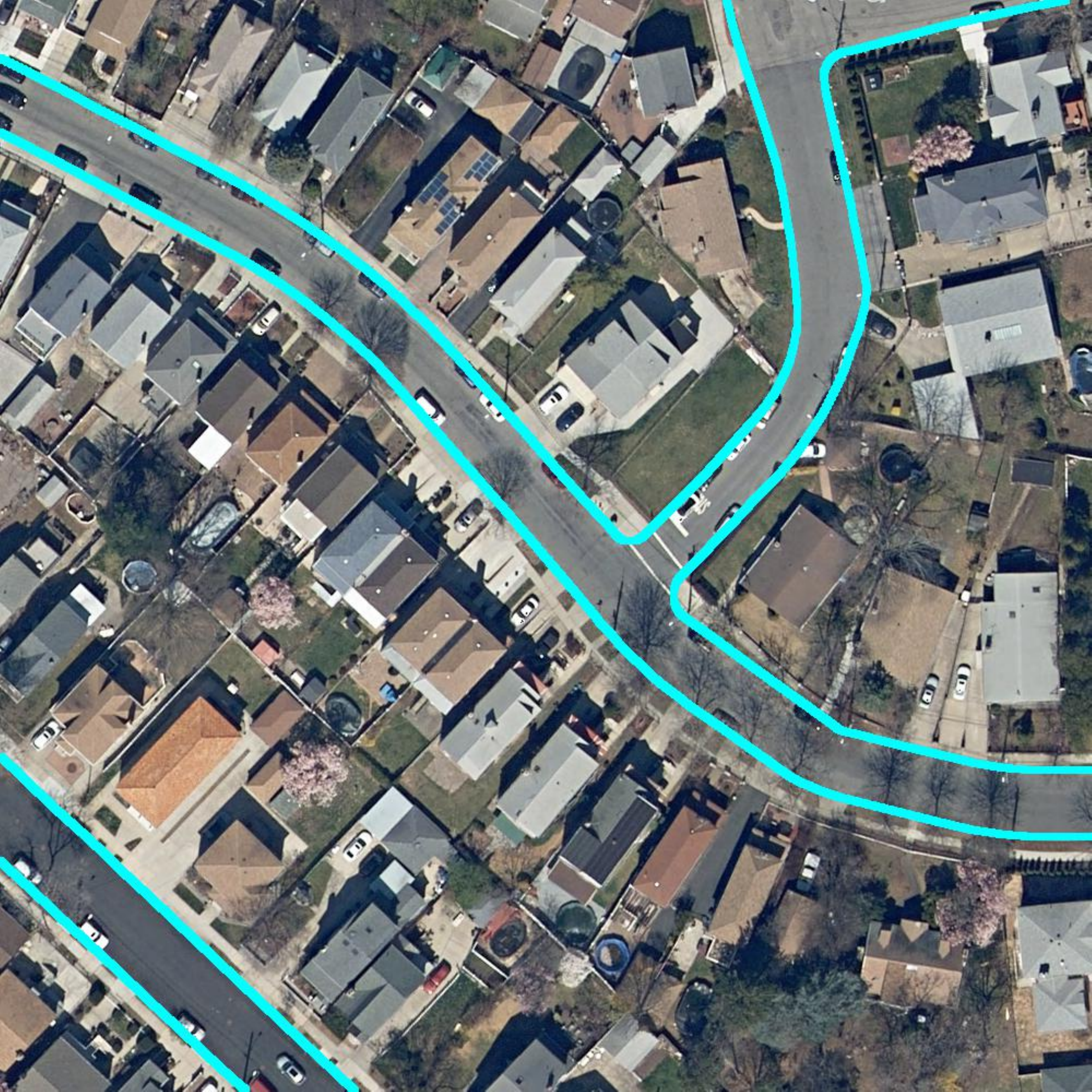}\caption{Ground-truth}
    \end{subfigure}\hfill
    \begin{subfigure}{.24\textwidth}
        \includegraphics[width=\textwidth]{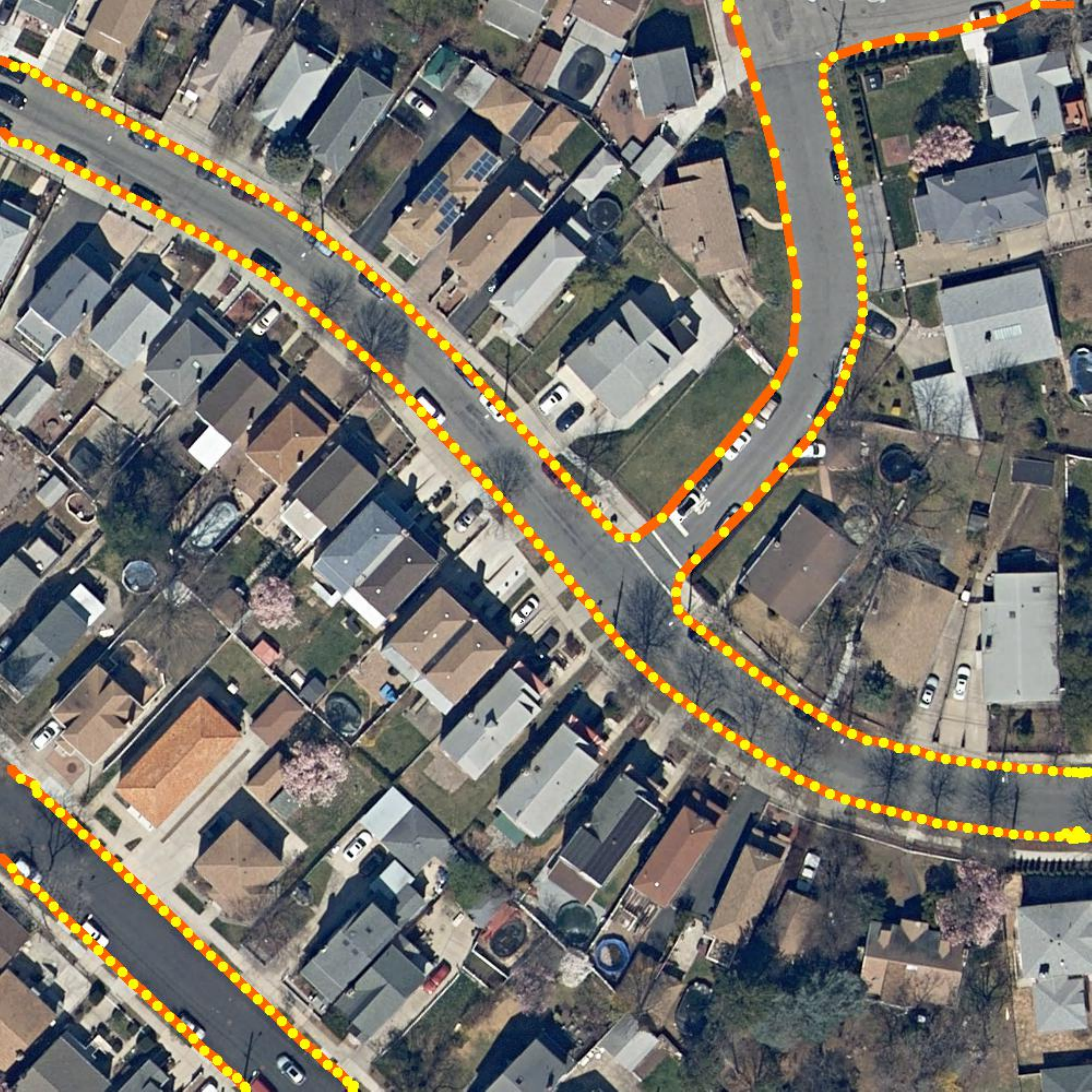}\caption{Our result}
    \end{subfigure}
    \caption{A sample result of our method. (a) The ground-truth road curb graph (cyan lines). (b) Our result (orange lines for edges and yellow nodes for vertices). We can see that our result is accurate and topologically correct. For better visualization, the lines are drawn with a thicker width while it is actually one-pixel width. The figure is best viewed in color.}
    \label{sample_demo_result}
\end{figure}
%%%%%%%%%%%% Figure 1 %%%%%%%%%%

Currently, there exists very limited work on off-line road-curb detection using aerial images, but similar work can be found in the area of line-shaped object detection. They can be generally divided into two categories: methods using image segmentation followed by post-processing algorithms \cite{mnih2010learning,mattyus2017deeproadmapper,batra2019improved}, and methods based on end-to-end iterative graph generation \cite{homayounfar2018hierarchical,liang2019convolutional,homayounfar2019dagmapper,bastani2018roadtracer}. 
The former methods usually first employ an image segmentation algorithm to roughly segment the line-shaped objects, and then use hard-engineered rules to refine the results.
Due to the unavoidable errors from image segmentation, the overall detection performance is usually bounded by the segmentation performance. In addition, the post-processing algorithms are usually based on heuristic assumptions, which could not always be satisfied in real traffic environments. To address these issues, the latter methods resort to end-to-end learning of line-shaped objects. They formulate the problem as an iterative graph generation problem. The input is the raw sensor measurement, and the output is directly the graph for the line-shaped objects. This paper adopts a similar paradigm as the latter one. It is worth noting that some works on road network extraction look similar to ours \cite{bastani2018roadtracer,tan2020vecroad}. Our work differs from them mainly in that they usually coarsely detect the central line of a road, but our work aims to accurately detect both sides of a road.

Imitation learning has drawn increasing attentions in recent years and has been used in many robotic applications, such as visual navigation \cite{cai2019vision} and robot control \cite{Zhu-RSS-18}. 
The basic idea of imitation learning is to mimic an expert control policy $\pi^*$ with a learned policy $\hat{\pi}$.  
In this paper, we solve the road-curb detection problem using imitation learning, and develop a novel end-to-end imitation learning network named as iCurb. Suppose we have an agent $\bm{A}$ locating at a vertex $v_t$ in an aerial image, $\bm{A}$ is expected to take an action controlled by $\hat{\pi}$ to move to the next vertex $v_{t+1}$ along road curbs based on the visual information near $v_t$. The expert demonstrations from $\pi^*$ can be generated from ground-truth road-curb labels. Our network aims to learn a policy $\hat{\pi}$ to approximate $\pi^*$, which controls the agent $\bm{A}$ to travel along road curbs from the initial vertex to the end vertex. In this way, the agent $\bm{A}$ can learn to \textit{draw} graphs representing road curbs vertex by vertex. 
To the best of our knowledge, this is the first work that detects road curbs using imitation learning. Fig. \ref{sample_demo_result} displays a sample result of our network. 
The contributions of this work are mainly three-fold:
\begin{enumerate}
    \item We innovatively propose an imitation learning-based solution for off-line road-curb detection.
    \item We design an agent network that can predict actions for the graph growing task.
    \item We propose a novel training strategy that consists of two exploration methods and a dynamic labeling method.
\end{enumerate}

%%%%%%%%%%%%%%%%%%%%%%%%%%%%%% Related work %%%%%%%%%%%%%%%%%%%%%%%%%%%%%%%%%%%%%
\section{Related Work}
\subsection{Automatic annotation by graph growing}
Manually annotating objects is tedious and time-consuming. Inspired by how humans annotate objects, Castrejon \textit{et al.} \cite{castrejon2017annotating} proposed the first algorithm to semi-automatically annotate object instances in images by growing a graph representing boundaries of target objects. This method was further improved by Acuna \textit{et al.} \cite{acuna2018efficient} which optimized the network structure and training strategy. Some works in autonomous driving \cite{homayounfar2018hierarchical,liang2019convolutional,homayounfar2019dagmapper} focused on automatically annotating line-shaped objects. For example, Homayounfar \textit{et al.} \cite{homayounfar2018hierarchical} detected lanes by first obtaining the initial vertices of each lane instances by a recurrent \textit{counting} head, and then training another recurrent \textit{drawing} head to grow each lane instance. Since this method cannot handle lanes with complex topological structures, such as forks or merges, they added another head to predict the merge or fork actions \cite{homayounfar2019dagmapper}. 

\subsection{Road network extraction}
With the advancement of deep-learning technologies, recent works on road network extraction resort to using convolutional neural network-based semantic segmentation. Mattyus \textit{et al.} \cite{mattyus2017deeproadmapper} proposed DeepRoadMapper, which could generate a road graph from rough discontinuous segmentation results by implementing a series of post-processing algorithms. But the underlying assumptions of the heuristic post-processing algorithms limited the method to be extended in more general scenarios. 

Instead of extracting road graph from segmentation results, some recent solutions directly generate a road graph from input aerial images \cite{bastani2018roadtracer,li2018polymapper,belli2019image,tan2020vecroad}. RoadTracer \cite{bastani2018roadtracer} proposed the first method to generate road graphs by iterative graph growing. Starting from a manually defined initial vertex, they iteratively predicted the next vertex and finally obtained the whole road graph. Based on this method, \cite{li2018polymapper,belli2019image,tan2020vecroad} further enhanced the performance by adopting better network structures and training strategies.

\subsection{Imitation leaning}
Imitation learning aims to learn a policy from expert demonstrations. 
%Usually, imitation learning requires a large amount of labeled training data so that the learned policy could cover a large state space. 
%In robotic applications, the training data mainly come from expert demonstrations, which may be not easy to collect. 
% on-robot \cite{rhinehart2018deep} or human expert \cite{stadie2017third} demonstrations,
Different from previous works, we generate labeled training data (expert demonstrations) from the ground truth of road curbs. 
Actually, methods proposed in past working on automatic annotation and road network extraction could be seen as the naive behavior cloning \cite{osa2018algorithmic}, which is a type of imitation learning algorithm. However, even though the learner could mimic the expert under most states, the learner tends to fail when exceptions occur since it does not learn how to recover \cite{ross2010efficient}. To address this problem, Ross \textit{et al.} \cite{ross2011reduction} proposed a meta algorithm called Dataset Aggregation (DAgger) to collect learner's behaviors into a dataset. In this way, DAgger can cover much larger state space so that it is able to handle exceptions. We propose our method based on DAgger.

%%%%%%%%%%%% Figure 2 %%%%%%%%%%
\begin{figure*}[t]
  \centering
    \includegraphics[width=.85\linewidth]{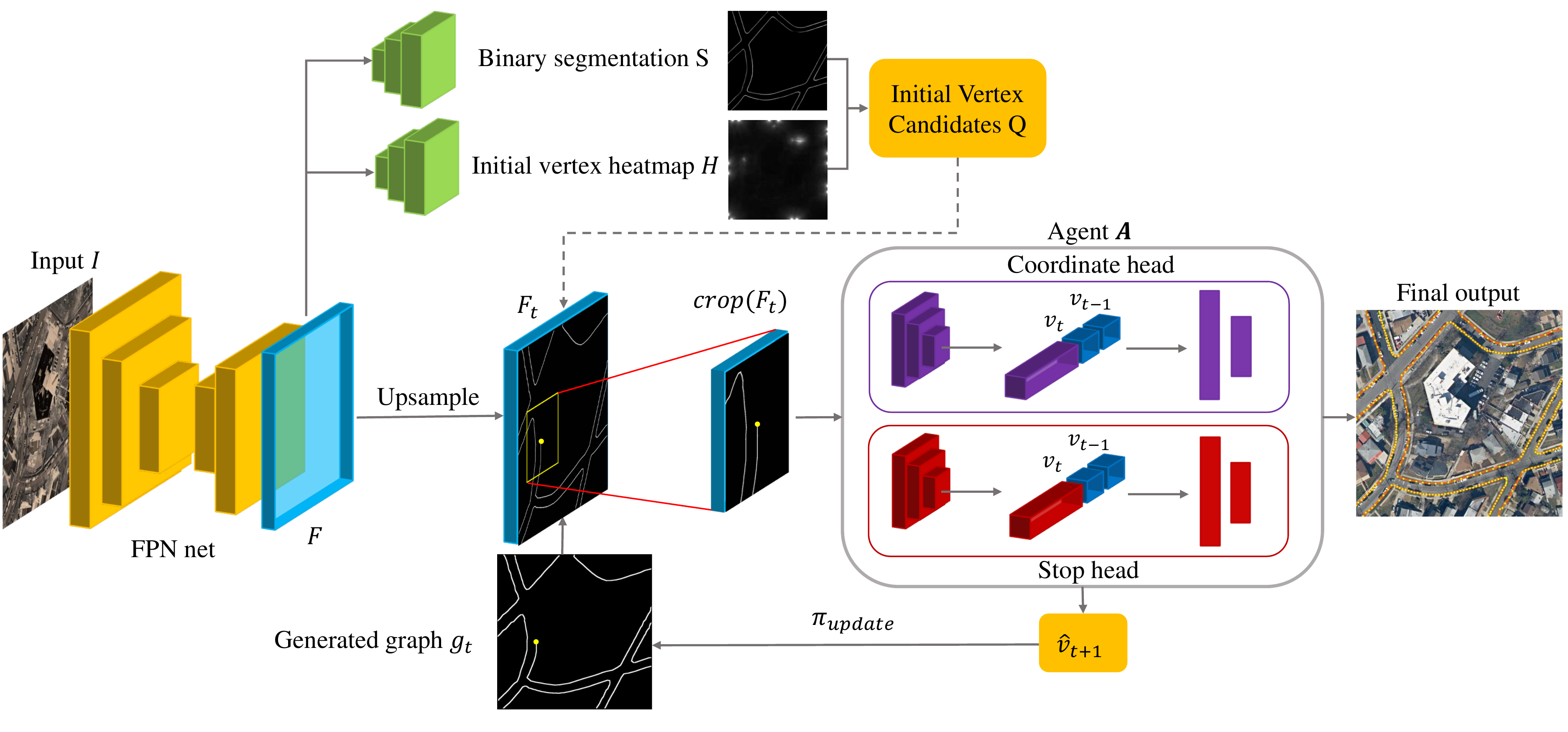}
  \caption{The overview diagram of our iCurb. We first extract the feature map $F$ by using the FPN network \cite{lin2017feature}. Two segmentation networks are trained to generate a set of initial vertex candidates $Q$ (The dashed line indicates that $Q$ is only used at the beginning of each iteration to grow a curb instance). The binary map $g_t$ records historical information of the generated graph until time $t$. $g_t$ is concatenated with up-sampled $F$ into $F_t$. Centering at the current vertex $v_t$ (denoted with the yellow node), we crop a $d\times d$ local feature map $crop(F_t)$ and send it into the agent $\bm{A}$ to predict the next vertex $\hat{v}_{t+1}=\{\hat{x}_{t+1},\hat{s}_{t+1}\}$. The coordinate head outputs the 2-D coordinates $\hat{x}_{t+1}$ and the stop head predicts the stop action $\hat{s}_{t+1}$. Then $g_t$ is updated based on policy $\pi_{update}$. iCurb could be trained in an end-to-end manner. The figure is best viewed in color. Please zoom in for details.}
  \label{diagram}
\end{figure*}
%%%%%%%%%%%% Figure 2 %%%%%%%%%%

%%%%%%%%%%%%%%%%%%%%%%%%%%%%%%%%%%%% Methodology %%%%%%%%%%%%%%%%%%%%%%%%%%%%  
\section{The Proposed Method}
\subsection{The Method Overview}\label{overview}

Fig. \ref{diagram} shows the overview diagram of our iCurb. Our problem can be decomposed into two sub-problems: (1) how to find the initial vertices to start the iterative graph growing; (2) how to iteratively grow a road curb instance from an initial vertex. For the former, we design an algorithm that generates the candidates of initial vertex based on road-curb segmentation results $S$ and initial vertex heatmaps $H$. As the latter is similar to visual navigation tasks, inspired by the DAgger algorithm \cite{ross2011reduction}, we propose a novel imitation learning algorithm to predict high-quality road-curb graphs. The output of our method is a graph $G=(V,E)$ that represents road curbs. The vertex set $V$ consists of the iteratively generated vertices $v_t=\{x_t,s_t\}$, where $x_t$ represents the 2-D coordinates of $v_t$, and $s_t$ is a variable to control the stop action. Once $s_t$ becomes 1, the agent $\bm{A}$ stops growing the current curb instance and turns to grow another one from an unprocessed initial vertex in $Q=\{q_i\}_{i=1}^N$. The edges in the set $E$ are simply obtained by connecting neighboring vertices during graph growing. In our work, we densify the initial polyline labels as the ground-truth road curbs.
%%%%%%%%%%%%%%% Figure 3%%%%%%%%
\begin{figure}[t]
  \centering
    \includegraphics[width=\linewidth]{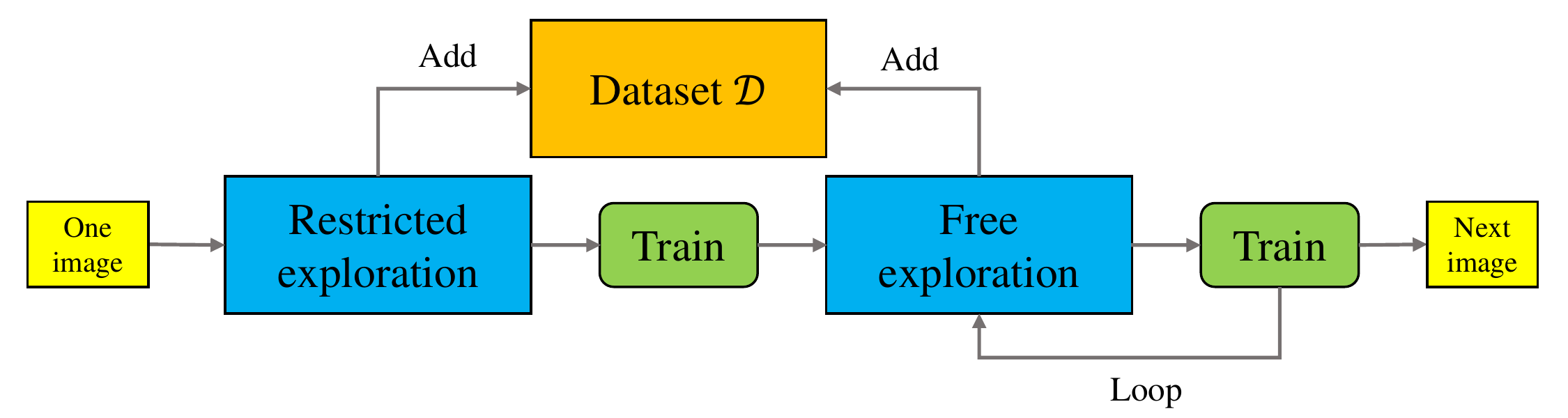}
  \caption{Diagram of our imitation learning algorithm based on DAgger. With an aerial image as input, iCurb first runs a restricted exploration, then several rounds of free explorations are implemented. After each exploration, generated samples are aggregated into dataset $\mathcal{D}$ and the agent $\bm{A}$ is trained on $\mathcal{D}$ once.}
  \label{DAgger_training}
\end{figure}
%%%%%%%%%%%%%%% Figure 3 %%%%%%%%%%

Our iCurb consists of three steps: 
Firstly, we extract the feature map $F$ using the Feature Pyramid Network (FPN) \cite{lin2017feature} from an input multi-channel aerial image $I$. We find the initial vertex candidates $Q=\{q_i\}_{i=1}^N$ by processing the binary segmentation map $S$ and initial vertex heatmap $H$. In order to save historical information, we set all the pixels covered by the predicted vertices or edges until time $t$ as 1 in $g_t$. 
Secondly, we concatenate $F$ and $g_t$ into a new multi-channel feature map $F_t$. Here, $F_t$ can be regarded as the environment in imitation learning. Assuming that the current vertex is $v_t$, we crop a $d\times d$ square block $crop(F_t)$ with the center at $v_t$ to represent the local feature of $v_t$. With $crop(F_t)$ given as input, the agent $\bm{A}$ makes the prediction for the next vertex $\hat{v}_{t+1}=\{\hat{x}_{t+1},\hat{s}_{t+1}\}$:
\begin{equation}
    \hat{v}_{t+1} = \argmax_{v_{t+1}\in crop(F_t)} {\pi[ v_{t+1}|crop(F_t),v_t,v_{t-1} ]},
\end{equation}
where $\pi$ is the policy of agent $\bm{A}$, $\hat{x}_{t+1}$ and $\hat{s}_{t+1}$ are predicted by the coordinate head and stop head, respectively. 
Finally, $g_t$ is updated given the policy $\pi_{update}$. During testing, $\pi_{update}$ directly adds the predicted $\hat{v}_{t+1}$ into $g_t$. But during training, the graph is grown with more complex strategies in order to generate training samples to teach the agent. The detailed training strategy is described in section \ref{imitation}. After all the initial vertex candidates in $Q$ being processed, the agent $\bm{A}$ is trained on $\mathcal{D}$ and the policy $\pi$ is updated based on the equation:
\begin{equation}
    \hat{\pi} = \argmin_{\pi}{\sum^{t}|\pi[crop(F_t),v_t,v_{t-1}]-v^*_{t+1}|}.
\end{equation}

\subsection{Feature extraction backbone}
We choose FPN as our backbone because it has been widely used to fuse multi-scale features for visual perception. In our task, road curbs are usually long and thin, so the backbone is expected to have a large receptive field and be able to capture detailed local spatial information.
Given as input a $4\times H\times W$ aerial image $I$, the FPN network generates an $8\times H/2\times W/2$ feature map $F$. $F$ is up-sampled by $4$ times to generate a feature map that has the original input resolution for graph growing. 
%%%%%%%%%%%% Algorithm 1  %%%%%%%%%%
\begin{algorithm}[t]
\KwInput{Input aerial images $X=\{I_i\}_{i=1}^M$}
\Begin{
Randomly initialize $\pi$ of agent $\bm{A}$ \\
\While{$X$ not empty}{
        $I\gets X.pop()$\\
        Initialize $\mathcal{D}\gets \emptyset$\\
        $\mathcal{D}_0 \gets$ restricted\_exploration(I)\\
        $\mathcal{D}\gets\mathcal{D}\cup\mathcal{D}_0$\\
        Train $\pi$ on $\mathcal{D}$\\
        \For{$i\gets1$ \KwTo $N$}{
              $\mathcal{D}_i \gets$ free\_exploration(I)\\
              $\mathcal{D}\gets\mathcal{D}\cup\mathcal{D}_i$\\
              Train $\pi$ on $\mathcal{D}$\\
       }
   }
}
\caption{The proposed DAgger-based algorithm}
\label{alg1}
\end{algorithm}
%%%%%%%%%%%% Algorithm 1  %%%%%%%%%%
\subsection{Segmentation and initial vertex candidates}
The two segmentation heads in our model aim to locate initial vertex candidates. One head outputs the binary segmentation probability map $S$ for road curbs. We first skeletonize\footnote{\url{https://scikit-image.org/docs/dev/auto_examples/edges/plot_skeleton.html}} $S$, then filter out short skeleton segments, and finally select one end vertex of each remained skeleton segments as the initial vertex candidates $Q$. Since $S$ may contain errors, we design the other segmentation head to predict the heatmap $H$ of the initial vertices. $H$ represents the probability distribution of the initial vertices. By calculating the local maximum of $H$, we could obtain another set of initial vertex candidates $Q'$. Then we add candidates in $Q'$ with high probability into $Q$ and remove those points in $Q$ whose probabilities in $H$ are low.
\subsection{Agent network}
In iCurb, the agent $\bm{A}$ consists of two prediction heads: one predicts the 2-D coordinates $x_{t+1}$ and the other one predicts the stop action $s_{t+1}$. They have the same network structure except for the number of output channels.
\subsubsection{Coordinate head}
The coordinate head takes as input $crop(F_t)$, and produces a feature vector that represents the local spatial information of the current vertex $v_t$. Then, it predicts a relative displacement $\Delta x_t$ of the current vertex within the scope of $crop(F_t)$. The vertex coordinate is updated by $x_{t+1}=x_t+\Delta x_t$. During training, $\Delta x_t$ is scaled from $[0,d]$ to $[-1,1]$. To enhance historical information and encode absolute coordinates, we concatenate past vertices $v_t$ and $v_{t-1}$ with the feature vector. Otherwise, $\bm{A}$ cannot know its absolute location in the whole image, which could degrade the final performance.
\subsubsection{Stop head} 
The stop head outputs 1 to trigger a stop action when one of the following events occurs: (1) The agent $\bm{A}$ reaches the end vertex of a curb instance, e.g., $\bm{A}$ reaches the end vertex on the edge of $I$; (2) $\bm{A}$ grows incorrect graphs, e.g., $\bm{A}$ generates vertices far from the ground-truth road curbs; (3) $\bm{A}$ spends time longer than a threshold on growing a curb instance, which could prevent local infinite loops. When a stop action is triggered, iCurb starts to grow another curb instance if there are vertices remained in $Q$. Otherwise, iCurb outputs the final graph for the image. Stop actions are rarer than normal actions, thus making the training data unbalanced. Therefore, with naive graph growing strategies proposed in previous works \cite{homayounfar2018hierarchical,homayounfar2019dagmapper,bastani2018roadtracer}, some stop actions may be missed in $\mathcal{D}$, and it will be hard for the agent $\bm{A}$ to handle possible errors during graph growing.
 
%%%%%%%%%%%% Algorithm 2  %%%%%%%%%%
\begin{algorithm}[t]
\KwInput{An aerial image $I$}
\KwOutput{Training samples of the aerial image $\mathcal{D}_I$}
\Begin{
Initialize initial vertex set $Q=\{q_i\}_{i=1}^N$\\
\While{$Q$ not empty}{
 $t\gets 0$ \\
 $v_t\gets Q.pop()$ \\
 \While{$s_t \neq 1$}{
 $v_{t+1}=\argmax{\pi[ v_{t+1}|crop(F_t),v_t,v_{t-1} ]}$\\
 $g_{t+1}\gets \pi_{update}(g_t,v_{t+1})$\\
 $F_{t+1} = concat(F,g_{t+1})$\\
 $\mathcal{D}_I\gets \mathcal{D}_I\cup \{crop(F_{t+1})\}$\\
 $t\gets t+1$
}
}
\Return $\mathcal{D}_I$
}
\caption{Exploration method}
\label{alg2}
\end{algorithm}
%%%%%%%%%%%% Algorithm 2  %%%%%%%%%%

%%%%%%%%%%%% Figure 4 %%%%%%%%%%%%%%%%%
 \begin{figure}[h]
 \centering
    \includegraphics[width=0.3\textwidth]{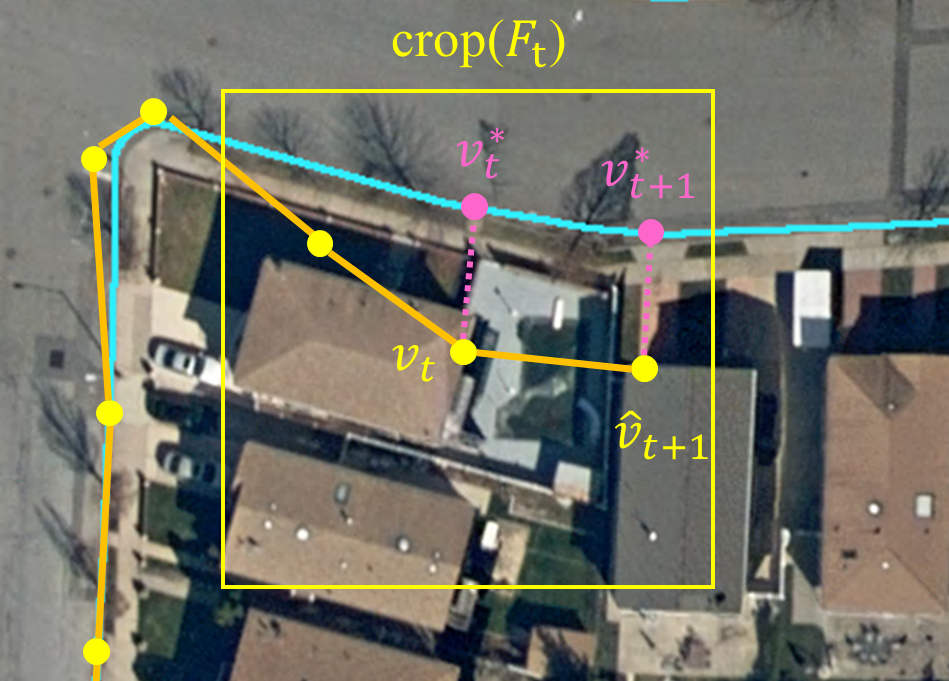}
    \caption{Visualization of how the restricted exploration method obtains $v_{t+1}^*$. The ground-truth road curbs are denoted by cyan lines. Yellow nodes represent the generated vertices and orange lines represent the generated edges. Suppose $\bm{A}$ is now at the vertex $v_t$ whose ground-truth label is $v^*_t$. Based on $crop(F_t)$ (represented by the yellow rectangle), the agent $\bm{A}$ makes the prediction of the next vertex $\hat{v}_{t+1}$. Then we find the point of ground-truth road curbs that is nearest to $\hat{v}_{t+1}$ as the label $v^*_{t+1}$. There are some restrictions on $v^*_{t+1}$, e.g., it must be after $v_t^*$ and the distance between them should be larger than a threshold. Please refer to the supplementary document for more details.}
    \label{res_exp}
\end{figure}
%%%%%%%%%%%%%%%% Figure 4 %%%%%%%%%%%%%%%%%%%%%%%%%

\subsection{Training strategy}\label{imitation}
Different from past works that apply either naive behavior cloning \cite{bastani2018roadtracer} or greedy algorithm \cite{homayounfar2018hierarchical,homayounfar2019dagmapper}, we propose a DAgger-based algorithm to teach the agent $\bm{A}$ how to mimic an expert policy.
The algorithm is described in Alg. \ref{alg1} and visualized in Fig. \ref{DAgger_training}. It has two methods to generate training samples: the restricted exploration method and the free exploration method. They share a common framework that is described in Alg. \ref{alg2}, but with a different update policy $\pi_{update}$. Please refer to the supplementary document for more details about our algorithm.
\subsubsection{Restricted exploration}
As the name suggests, the restricted exploration method generates training samples under some restrictions. For example, the agent $\bm{A}$ should not be far away from the ground-truth road curbs. So for each prediction $\hat{v}_{t+1}$, we find the nearest point $v^*_{t+1}$ in the ground-truth road curbs using the method described in Fig. \ref{res_exp}. 
When the distance is smaller than a threshold (i.e., 15 pixels in our paper), we directly update the graph $g_t$ by adding $\hat{v}_{t+1}$ into it. But when the distance is larger than the threshold, we add $v^*_{t+1}$ instead of $\hat{v}_{t+1}$ into $g_t$. In this way, we are able to keep $\bm{A}$ moving near the right track and more training samples around the ground-truth road curbs can be generated. 
\subsubsection{Free exploration}
The restricted exploration has an issue that it decreases the opportunities to cover more stop action events, because the movement of $\bm{A}$ is restricted within 15 pixels around the ground-truth road curb. Moreover, since there is no restriction for $\bm{A}$ during testing, a data distribution gap between training and testing would occur. To alleviate these issues, we propose the free exploration method for $\bm{A}$, in which the update policy $\pi_{update}$ directly adds $\hat{v}_{t+1}$ into $g_t$.
\subsubsection{Combination}
The restricted exploration method generates training samples around the ground-truth road curbs, which could accelerate convergence and improve the accuracy of obtained graphs, but it would make $\bm{A}$ vulnerable to errors. On the contrary, the free exploration method could enhance the robustness of $\bm{A}$, but make the training data noisier and even significantly slow down the convergence. So we combine them together and propose a training strategy based on DAgger, which is shown in Alg. \ref{alg1} and Fig. \ref{DAgger_training}.

%%%%%%%%%%%% Figure 5  %%%%%%%%%%
 \begin{figure*}[t]
\newcommand{\picvis}[1]{002187#144}
    \newcommand{\picvisi}[1]{002185#110}
    \newcommand{\picvisii}[1]{005225#144}
    \newcommand{\picvisiii}[1]{012200#133}
    \newcommand{\picvisiv}[1]{025265#120}
    
    \newcommand{\picvisv}[1]{025250#100}
    \newcommand{\picvisvi}[1]{937165#110}
    \newcommand{\picvisvii}[1]{015232#113}
 \centering
    \begin{subfigure}[t]{0.105\textwidth}
        \begin{subfigure}[t]{\textwidth}
            \includegraphics[width=\textwidth]{gt_\picvisi{_}.pdf}
        \end{subfigure}\vspace{.6ex}
        \begin{subfigure}[t]{\textwidth}
            \includegraphics[width=\textwidth]{gt_\picvisii{_}.pdf}
        \end{subfigure}\vspace{.6ex}
        \begin{subfigure}[t]{\textwidth}
            \includegraphics[width=\textwidth]{gt_\picvisvii{_}.pdf}
        \end{subfigure}\vspace{.6ex}
        \begin{subfigure}[t]{\textwidth}
            \includegraphics[width=\textwidth]{gt_\picvisiv{_}.pdf}
        \end{subfigure}\vspace{.6ex}
        \caption{GT}
        \label{fig_qualitative_1st}
    \end{subfigure}
    %naive
    \begin{subfigure}[t]{0.105\textwidth}
        \begin{subfigure}[t]{\textwidth}
            \includegraphics[width=\textwidth]{naive_\picvisi{_}.pdf}
        \end{subfigure}\vspace{.6ex}
        \begin{subfigure}[t]{\textwidth}
            \includegraphics[width=\textwidth]{naive_\picvisii{_}.pdf}
        \end{subfigure}\vspace{.6ex}
        \begin{subfigure}[t]{\textwidth}
            \includegraphics[width=\textwidth]{naive_\picvisvii{_}.pdf}
        \end{subfigure}\vspace{.6ex}
        \begin{subfigure}[t]{\textwidth}
            \includegraphics[width=\textwidth]{naive_\picvisiv{_}.pdf}
        \end{subfigure}\vspace{.6ex}
        \caption{Naive}
        \label{fig_qualitative_1st}
    \end{subfigure}
    % learning
    \begin{subfigure}[t]{0.105\textwidth}
        \begin{subfigure}[t]{\textwidth}
            \includegraphics[width=\textwidth]{lb_\picvisi{_}.pdf}
        \end{subfigure}\vspace{.6ex}
        \begin{subfigure}[t]{\textwidth}
            \includegraphics[width=\textwidth]{lb_\picvisii{_}.pdf}
        \end{subfigure}\vspace{.6ex}
        \begin{subfigure}[t]{\textwidth}
            \includegraphics[width=\textwidth]{lb_\picvisvii{_}.pdf}
        \end{subfigure}\vspace{.6ex}
        \begin{subfigure}[t]{\textwidth}
            \includegraphics[width=\textwidth]{lb_\picvisiv{_}.pdf}
        \end{subfigure}\vspace{.6ex}
        \caption{\cite{mnih2010learning}}
        \label{fig_qualitative_1st}
    \end{subfigure}
    % deep
    \begin{subfigure}[t]{0.105\textwidth}
        \begin{subfigure}[t]{\textwidth}
            \includegraphics[width=\textwidth]{deep_\picvisi{_}.pdf}
        \end{subfigure}\vspace{.6ex}
        \begin{subfigure}[t]{\textwidth}
            \includegraphics[width=\textwidth]{deep_\picvisii{_}.pdf}
        \end{subfigure}\vspace{.6ex}
        \begin{subfigure}[t]{\textwidth}
            \includegraphics[width=\textwidth]{deep_\picvisvii{_}.pdf}
        \end{subfigure}\vspace{.6ex}
        \begin{subfigure}[t]{\textwidth}
            \includegraphics[width=\textwidth]{deep_\picvisiv{_}.pdf}
        \end{subfigure}\vspace{.6ex}
        \caption{\cite{mattyus2017deeproadmapper}}
        \label{fig_qualitative_1st}
    \end{subfigure}
    % refine
    \begin{subfigure}[t]{0.105\textwidth}
        \begin{subfigure}[t]{\textwidth}
            \includegraphics[width=\textwidth]{refine_\picvisi{_}.pdf}
        \end{subfigure}\vspace{.6ex}
        \begin{subfigure}[t]{\textwidth}
            \includegraphics[width=\textwidth]{refine_\picvisii{_}.pdf}
        \end{subfigure}\vspace{.6ex}
        \begin{subfigure}[t]{\textwidth}
            \includegraphics[width=\textwidth]{refine_\picvisvii{_}.pdf}
        \end{subfigure}\vspace{.6ex}
        \begin{subfigure}[t]{\textwidth}
            \includegraphics[width=\textwidth]{refine_\picvisiv{_}.pdf}
        \end{subfigure}\vspace{.6ex}
        \caption{\cite{batra2019improved}}
        \label{fig_qualitative_1st}
    \end{subfigure}
    % roadtracer
    \begin{subfigure}[t]{0.105\textwidth}
        \begin{subfigure}[t]{\textwidth}
            \includegraphics[width=\textwidth]{rt_\picvisi{_}.pdf}
        \end{subfigure}\vspace{.6ex}
        \begin{subfigure}[t]{\textwidth}
            \includegraphics[width=\textwidth]{rt_\picvisii{_}.pdf}
        \end{subfigure}\vspace{.6ex}
        \begin{subfigure}[t]{\textwidth}
            \includegraphics[width=\textwidth]{rt_\picvisvii{_}.pdf}
        \end{subfigure}\vspace{.6ex}
        \begin{subfigure}[t]{\textwidth}
            \includegraphics[width=\textwidth]{rt_\picvisiv{_}.pdf}
        \end{subfigure}\vspace{.6ex}
        \caption{\cite{bastani2018roadtracer}}
        \label{fig_qualitative_1st}
    \end{subfigure}
    % vecroad
    \begin{subfigure}[t]{0.105\textwidth}
        \begin{subfigure}[t]{\textwidth}
            \includegraphics[width=\textwidth]{vr_\picvisi{_}.pdf}
        \end{subfigure}\vspace{.6ex}
        \begin{subfigure}[t]{\textwidth}
            \includegraphics[width=\textwidth]{vr_\picvisii{_}.pdf}
        \end{subfigure}\vspace{.6ex}
        \begin{subfigure}[t]{\textwidth}
            \includegraphics[width=\textwidth]{vr_\picvisvii{_}.pdf}
        \end{subfigure}\vspace{.6ex}
        \begin{subfigure}[t]{\textwidth}
            \includegraphics[width=\textwidth]{vr_\picvisiv{_}.pdf}
        \end{subfigure}\vspace{.6ex}
        \caption{\cite{tan2020vecroad}}
        \label{fig_qualitative_1st}
    \end{subfigure}
    % dagmapper
    \begin{subfigure}[t]{0.105\textwidth}
        \begin{subfigure}[t]{\textwidth}
            \includegraphics[width=\textwidth]{dm_\picvisi{_}.pdf}
        \end{subfigure}\vspace{.6ex}
        \begin{subfigure}[t]{\textwidth}
            \includegraphics[width=\textwidth]{dm_\picvisii{_}.pdf}
        \end{subfigure}\vspace{.6ex}
        \begin{subfigure}[t]{\textwidth}
            \includegraphics[width=\textwidth]{dm_\picvisvii{_}.pdf}
        \end{subfigure}\vspace{.6ex}
        \begin{subfigure}[t]{\textwidth}
            \includegraphics[width=\textwidth]{dm_\picvisiv{_}.pdf}
        \end{subfigure}\vspace{.6ex}
        \caption{\cite{homayounfar2019dagmapper}}
        \label{fig_qualitative_1st}
    \end{subfigure}
    % icurb
    \begin{subfigure}[t]{0.105\textwidth}
        \begin{subfigure}[t]{\textwidth}
            \includegraphics[width=\textwidth]{icurb_\picvisi{_}.pdf}
        \end{subfigure}\vspace{.6ex}
        \begin{subfigure}[t]{\textwidth}
            \includegraphics[width=\textwidth]{icurb_\picvisii{_}.pdf}
        \end{subfigure}\vspace{.6ex}
        \begin{subfigure}[t]{\textwidth}
            \includegraphics[width=\textwidth]{icurb_\picvisvii{_}.pdf}
        \end{subfigure}\vspace{.6ex}
        \begin{subfigure}[t]{\textwidth}
            \includegraphics[width=\textwidth]{icurb_\picvisiv{_}.pdf}
        \end{subfigure}\vspace{.6ex}
        \caption{Ours}
        \label{fig_qualitative_1st}
    \end{subfigure}

    \caption{Qualitative demonstrations. Each row shows an example. The first column is the ground-truth (cyan lines); column (b) to (e) are segmentation-based baselines (green lines); column (f) to (h) are modified graph-based baselines (pink lines for edges and yellow nodes for vertices); the last column shows the results of our iCurb (orange lines for edges and yellow nodes for vertices). Compared with baselines, the results of iCurb are more accurate and topologically correct. For better visualization, the lines are drawn in a thicker width while they are actually one-pixel width. The figure is best viewed in color. Please zoom in for details.}
    \label{fig_qualitative}
\end{figure*}
 %%%%%%%%%%%% Figure 5  %%%%%%%%%%
\subsubsection{Dynamic labels}
In our experiments, we find that it is hard to correctly find the training label $v^*_{t+1}$ merely based on $crop(F_t)$, because $v^*_{t+1}$ is acceptable as long as it lies on the unexplored ground-truth road-curb graph. So we cannot simply generate $v^*_{t+1}$ when $\bm{A}$ generates training samples. In this work, we find $v^*_{t+1}$ only after the prediction $\hat{v}_{t+1}$ is obtained. Specifically, for each training sample, we find the nearest point to $\hat{v}_{t+1}$ in the ground-truth road curbs within $crop(F_t)$ as $v^*_{t+1}$. In this way, the training sample becomes more reasonable. In short, different from the original DAgger algorithm that adds the current state and the corresponding expert action $[s,\pi^*(s)]$ into $\mathcal{D}$, we dynamically generate the training label (i.e., $\pi^*(s)=v^*_{t+1}$) on-the-fly based on the prediction $\hat{v}_{t+1}$.

%%%%%%%%%%%%%%%%%%%%%%%%% Experiment %%%%%%%%%%%%%%%%%%%%%%%%%%%%%%%%%%%
\section{Experimental Results and Discussions}
\subsection{Dataset}
In the past, there are very few large-scale datasets for road-curb detection, which hinders the learning-based approaches. Recently, NYC OpenData (New York City OpenData) added road curbs as a new feature into the NYC-planimetrics dataset \cite{nyc_dataset}. We then create our own dataset by pre-processing \cite{nyc_dataset}. There are 2,049 4-channel $5000\times 5000$ high-resolution aerial images in the dataset. The 4 channels are red, green, blue, and infrared. The images have 0.5 ft/pixel $\approx$ 15.2 cm/pixel resolution.  

To facilitate our training, we split each image and its annotations into 25 $1000\times1000$ smaller patches. Then we remove the patches without road curbs or inappropriate annotations (e.g., incorrect annotation). We finally get a pre-processed dataset that has 29,000 $1000\times 1000$ images. Among them, 18,000 images are used to pre-train the segmentation networks. With the pre-trained weights, our iCurb can convergence with less time. We use another 10,000 images to train our iCurb and the remaining 1,000 images for testing.

\subsection{Training Process}
As aforementioned, we first pre-train the segmentation networks. Then, during the training process of our iCurb, we load the pre-trained weights for the segmentation networks, and fix the model parameters. The algorithm for training our iCurb is described in Alg. \ref{alg1} and Fig. \ref{DAgger_training}. 
The training process usually takes a long time, because for each image $\bm{A}$ runs one time of restricted exploration and $N$ times of free exploration ($N=5$ in our experiment). 
During training, the initial vertex candidates $Q$ are obtained by adding Gaussian noises to the ground-truth initial vertices, while during testing, $Q$ is generated from the segmentation results $S$ and $H$ by the proposed algorithm. The experiments are conducted with a single GTX1080Ti GPU and a Intel i7-8700K CPU.

\subsection{Evaluation metrics}
We evaluate our results using pixel-wise metrics (i.e., Precision, Recall, and F1-score) and the Customized Connectivity ($CC$) metric inspired by \cite{homayounfar2018hierarchical}.

Suppose $P_{GT}$ is the set of pixels covered by ground-truth graphs and $P_{pre}$ is the set of pixels covered by predicted graphs. Then we define the Precision as the number of pixels in $P_{pre}$ that fall within a threshold $\tau$ to the ground-truth graph divided by the number of pixels in $P_{pre}$. Similarly, Recall is obtained by calculating the number of pixels in $P_{GT}$ that fall within $\tau$ to the predicted graph divided by the number of pixels in $P_{GT}$. 
The formulas for the metrics are:
\begin{equation} 
\begin{aligned}
        &P = \frac{|\{
        p | d(p,P_{GT})<\tau,\forall p\in P_{pre}
        \}|}{|P_{pre}|},\\
        &R = \frac{|\{
        p | d(p,P_{pre})<\tau,\forall p\in P_{GT}
        \}|}{|P_{GT}|},\\
        &F1\text{-}score =\frac{2P\cdot R}{P+R},
\end{aligned}
\end{equation}
where function $d(a,B)$ calculates the euclidean distance from a point $a$ to the closest point in a set $B$, $\{\cdot\}$ represents a set, and $|\cdot|$ represents the number of elements in a set. $P$ is Precision and $R$ is Recall. In this paper, we report the experimental results with $\tau$ set to 1.0, 2.0, 5.0 and 10.0 pixels, respectively. Larger $\tau$ would lead to better results, because it gives more tolerance to the prediction.

The metric $CC$ is calculated by:
\begin{equation}
CC = \sum_{i=1}^N K_i, \\ 
K_i = \left\{
             \begin{aligned}
          \frac{\alpha_i}{m_i}, &m_i\neq 0\\
          0 , &m_i = 0
        \end{aligned}
\right.
\end{equation}
where $K_i$ indicates the connectivity of the $i$-th ground-truth curb instance and $N$ is the number of ground-truth curb instances. 
For each predicted curb instance, we find the closest ground-truth instance (suppose it is the j-th ground-truth instance) to it and add $m_j$ by 1. Intuitively, $m_i$ represents the number of predicted curbs that are used to approximate the $i$-th ground-truth curb instance, and larger $m_i$ means worse connectivity. Then, we multiply $1/m_i$ with a weighting coefficient $\alpha_i$, which is equal to the number of pixels covered by the $i$-th ground-truth curb instance divided by the total number of pixels covered by the ground-truth curb. For each metric above, larger values indicate better performance.

\subsection{Ablation study}

In this section, we study the significance of some components of our network and training strategy. The quantitative results are shown in Tab. \ref{tab_ablation}.

Firstly, we create a variant by removing historical vertices $v_t$ and $v_{t-1}$ from the feature vector in $\bm{A}$. Then $\bm{A}$ would make the prediction only based on the local feature $crop(F_t)$, so it has no information about past positions in the global image coordinate system. This invariant presents a much lower performance in terms of both F1-score and connectivity. Therefore, the historical vertices $v_t$ and $v_{t-1}$ are necessary in iCurb.

Secondly, we remove the restricted exploration method. From the results, we find that this variant presents slightly inferior results than the original iCurb. Because of the existence of the free exploration method, the agent $\bm{A}$ could still learn to handle exceptions and achieves relatively satisfactory results. However, without the restriction method, the training samples become noisy. So this variant could not have results as good as the original iCurb. In addition, the convergence is slowed down because of the noisy training data. The experimental results confirm the necessity of the restricted exploration in our method.

Thirdly, we remove the free exploration method from the original iCurb. From Tab. \ref{tab_ablation}, we find that removing the free exploration method severely degrades the results. As aforementioned, the free exploration method generates training samples covering larger state space so that $\bm{A}$ could learn to handle exceptions. Thus removing the free exploration method makes this variant unable to take correct stop actions. The Recall remains high because wrongly generated graphs cover larger areas. So high recall here does not indicate better performance. Therefore, the free exploration method is critical to our method.

  %%%%%%%%%%%% Table 1  %%%%%%%%%%
\begin{table*}[h] 
\setlength{\abovecaptionskip}{0pt} 
\setlength{\belowcaptionskip}{0pt} 
\renewcommand\arraystretch{1.0} %row height
\renewcommand\tabcolsep{7.8pt} 
\centering 
\begin{threeparttable}
\caption{The quantitative results for the ablation study. The best results are highlighted in bold font. For all the metrics, larger values indicate better performance. We assess the historical vertices $v_t$ and $v_{t-1}$ (V), the restricted exploration (R), the free exploration (F) and DAgger-based training strategy (D).} 
\begin{tabular}{c c c c c c c c c c c c c c c c c}
\toprule
\multicolumn{4}{c}{}& \multicolumn{4}{c}{Precision} & \multicolumn{4}{c}{Recall} & \multicolumn{4}{c}{F1-score} & \multirow{3}{*}{$CC$} \\ 
\cmidrule(l){5-8} \cmidrule(l){9-12} \cmidrule(l){13-16} 
V  & R & F & D & 1.0 &  2.0 &  5.0 &  10.0 &  1.0 &  2.0 &  5.0 &  10.0
&  1.0 &  2.0 &  5.0 &  10.0 & \\
\midrule
&\checkmark&\checkmark&\checkmark &0.114 & 0.386 & 0.816 & 0.930 & 0.126 & 0.410 & 0.832 & 0.929 & 0.118 & 0.394 & 0.818 & 0.923 & 0.836 \\
\checkmark& &\checkmark&\checkmark &0.151 & 0.497 & 0.869 & 0.929 & 0.158 & 0.508 & 0.873 & 0.926 & 0.154 & 0.499 & 0.865 & 0.921 & 0.854\\
\checkmark&\checkmark& &\checkmark &0.121 & 0.393 & 0.676 & 0.735 & 0.161 & 0.518 & 0.876 & 0.928 & 0.136 & 0.439 & 0.750 & 0.807 &0.739 \\
\checkmark&\checkmark&\checkmark& & 0.170 & 0.536 & 0.878 & 0.935 & 0.172 & 0.533 & 0.865 & 0.914 & 0.169 & 0.531 & 0.865 & 0.918 & \textbf{0.877} \\
\midrule
\checkmark&\checkmark&\checkmark&\checkmark & \textbf{0.173} & \textbf{0.554} & \textbf{0.887} & \textbf{0.938} & \textbf{0.179} & \textbf{0.565} &\textbf{0.890}& \textbf{0.934} & \textbf{0.175} & \textbf{0.556} & \textbf{0.883} & \textbf{0.930} & 0.866
 \\
\bottomrule 
\label{tab_ablation}
\end{tabular} 
\end{threeparttable}
\end{table*}
%%%%%%%%%%%% Table 1  %%%%%%%%%%
%%%%%%%%%%%% Table 2  %%%%%%%%%%
\begin{table*}[th] 
\setlength{\abovecaptionskip}{0pt} 
\setlength{\belowcaptionskip}{0pt} 
\renewcommand\arraystretch{1.0} %row height
\renewcommand\tabcolsep{7.4pt} 
\centering 
\begin{threeparttable}
\caption{The quantitative results for the comparison. The best results are highlighted in bold font. For all the metrics, larger values indicate better performance.} 
\begin{tabular}{c c c c c c c c c c c c c c c c c}
\toprule
\multirow{3}{*}{Methods}& \multicolumn{4}{c}{Precision} & \multicolumn{4}{c}{Recall} & \multicolumn{4}{c}{F1-score} & \multirow{3}{*}{$CC$} \\ 
\cmidrule(l){2-5} \cmidrule(l){6-9} \cmidrule(l){10-13} 
& 1.0 &  2.0 &  5.0 &  10.0 &  1.0 &  2.0 &  5.0 &  10.0
&  1.0 &  2.0 &  5.0 &  10.0 & \\
\midrule
Naive baseline & 0.178 & 0.556 & 0.854 & 0.890 & 0.164 & 0.512 & 0.783 & 0.820 & 0.166 & 0.517 & 0.793 & 0.829 & 0.629 \\ 
Iterative refinement \cite{mnih2010learning} & 0.161 & 0.556 & 0.829 & 0.847 & 0.201 & 0.511 & 0.776 & 0.809 & 0.163 & 0.515 & 0.783 & 0.811 & 0.641 \\ 
Deeproadmapper \cite{mattyus2017deeproadmapper} & 0.152 & 0.479 & 0.742 & 0.787 & 0.184 & 0.582 & 0.896 & \textbf{0.941} & 0.164 & 0.515 & 0.796 & 0.842 &
0.643 \\ 
Improved Connectivity \cite{batra2019improved} &\textbf{0.202} & \textbf{0.593} & 0.834 & 0.857 & \textbf{0.218} & \textbf{0.643} & \textbf{0.903} & 0.928 & \textbf{0.206} & \textbf{0.607} & 0.854 & 0.878 & 0.702\\ 
\midrule
RoadTracer \cite{bastani2018roadtracer} & 0.092 & 0.305 & 0.590 & 0.688 & 0.119 & 0.392 & 0.738 & 0.836 & 0.101 & 0.334 & 0.640 & 0.736 & 0.767
\\ 
VecRoad \cite{tan2020vecroad}& 0.108 & 0.368 & 0.716 & 0.813 & 0.110 & 0.372 & 0.720 & 0.808 & 0.107 & 0.365 & 0.708 & 0.799 & 0.820
\\ 
DAGMapper \cite{homayounfar2019dagmapper} & 0.081 & 0.283 & 0.650 & 0.830 & 0.084 & 0.291 & 0.662 & 0.832 & 0.082 & 0.283 & 0.648 & 0.821 & 0.832
\\ 
\midrule
iCurb & 0.173 & 0.554 & \textbf{0.887} & \textbf{0.938} & 0.179 & 0.565 & 0.890 & 0.934 & 0.175 & 0.556 & \textbf{0.883} & \textbf{0.930} & \textbf{0.866} \\
\bottomrule 
\label{tab_comparative}
\end{tabular} 
\end{threeparttable}
\end{table*}
%%%%%%%%%%%% Table 2 %%%%%%%%%%%%%%

%%%%%%%%%%%% Table 3  %%%%%%%%%%
\begin{table}[th] 
\setlength{\abovecaptionskip}{0pt} 
\setlength{\belowcaptionskip}{0pt} 
\renewcommand\arraystretch{1.0} %row height
\renewcommand\tabcolsep{15pt} 
\centering 
\begin{threeparttable}
\caption{The time cost for comparison. We report the average time taken to process a single image.} 
\begin{tabular}{c c c}
\toprule
& Training & Inference \\
\midrule
Naive baseline & 0.38s/image & 0.40s/image \\ 
Iterative refinement \cite{mnih2010learning} & 0.74s/image & 0.64s/image \\ 
Deeproadmapper \cite{mattyus2017deeproadmapper} & 2.70s/image & 2.43s/image\\ 
Improved Connectivity \cite{batra2019improved} & 1.90s/image & 1.36s/image \\ 
\midrule
RoadTracer \cite{bastani2018roadtracer} &5.52s/image & 0.88s/image \\ 
VecRoad \cite{tan2020vecroad} & 15.16s/image & 1.16s/image\\ 
DAGMapper \cite{homayounfar2019dagmapper} &9.46s/image & 0.76s/image\\ 
\midrule
iCurb & 22.38s/image & 0.89s/image  \\
\bottomrule 
\label{tab_time}
\end{tabular} 
\end{threeparttable}
\end{table}
%%%%%%%%%%%% Table 3
Finally, to justify the superiority of the DAgger-based algorithm, we modify the training strategy by cleaning $\mathcal{D}$ every time after training. Then the algorithm becomes more like SEARN \cite{daume2009search}, which only trains on the data of one iteration (i.e., $\mathcal{D}_i$) instead of the aggregated data (i.e., $\mathcal{D}$). Because most errors made by $\bm{A}$ occur in the early stage, the data aggregation could enhance the robustness of $\bm{A}$ by training early samples for multiple times. In Tab. \ref{tab_ablation}, we find that this variant has inferior results than the initial iCurb due to the lack of data aggregation. Thus DAgger-based algorithm is essential for iCurb.

\subsection{Comparative results}

To the best of our knowledge, there is currently no published work on road-curb detection from aerial images. 
So for the method comparison, we build baselines by ourselves, including four segmentation-based baselines and three graph-based baselines. The comparative results are shown in Tab. \ref{tab_comparative} and time analysis is shown in Tab. \ref{tab_time}. 

For segmentation-based methods, we first create a naive baseline by doing a series of processing to the segmentation results $S$, such as thresholding and skeletonization. This simple method presents a good F1-score since the semantic segmentation network gives a good initial guess. However, it has poor connectivity which is not acceptable for real applications. Then, we build three baselines that can relieve the disconnectivity issue based on past works \cite{mnih2010learning,mattyus2017deeproadmapper,batra2019improved}. Among them, \cite{mattyus2017deeproadmapper} can generate new connections to bridge the disconnections in the segmentation results, and \cite{mnih2010learning,batra2019improved} iteratively refine the segmentation results  with another deep network. However, even though these three baselines could improve the F1-score and connectivity to some extent, they are still pixel-wise and do not have enough constraints on the connectivity. Thus, the results are not sufficient for our task.

Different from segmentation-based methods, graph-based methods could better emphasize connectivity. We adapt three works \cite{bastani2018roadtracer,tan2020vecroad,homayounfar2019dagmapper} on line-shaped object detection, to make them applicable to our task. Among them, RoadTracer \cite{bastani2018roadtracer} has constant step length, while VecRoad \cite{tan2020vecroad} and DagMapper \cite{homayounfar2019dagmapper} have flexible step length. Even though the connectivity is greatly enhanced, none of these past works realize the similarity between graph-based methods and imitation learning, so they tend to have pretty straight and naive training strategies. As a result, they perform poorly on handling exceptions and cannot correctly trigger the stop action very well. Thus, they generate a lot of incorrect vertices and no corresponding corrections can be made, which yields much worse F1-score and qualitative results. Therefore, the superiority of the training strategy of iCurb is demonstrated.

Since graph-based methods are sequential and generate labels on-the-fly, they tend to take longer processing time than segmentation-based methods. During the training period, iCurb runs 1+N rounds of exploration for every image, so more time consumption is needed compared with other graph-based baselines. But for inference, the efficiency of graph-based methods is satisfactory. From Tab. \ref{tab_time}, we note that iCurb takes more training time, but the inference speed of our method is competitive compared with other methods.

\subsection{Failure cases}

\begin{figure}[t]
 \begin{subfigure}[t]{0.155\textwidth}
        \begin{subfigure}[t]{\textwidth}
            \includegraphics[width=\textwidth]{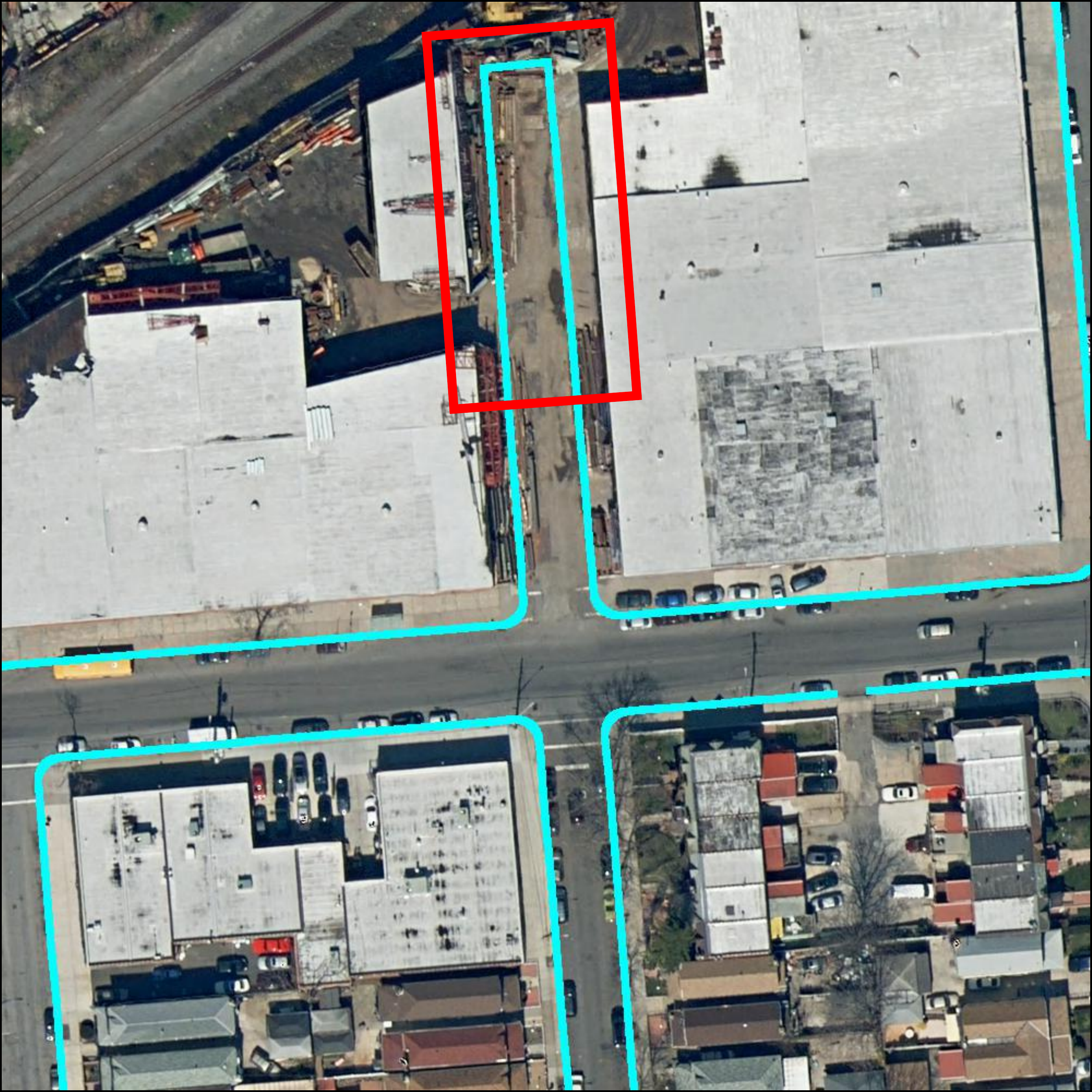}
        \end{subfigure}\vspace{.6ex}
        \begin{subfigure}[t]{\textwidth}
            \includegraphics[width=\textwidth]{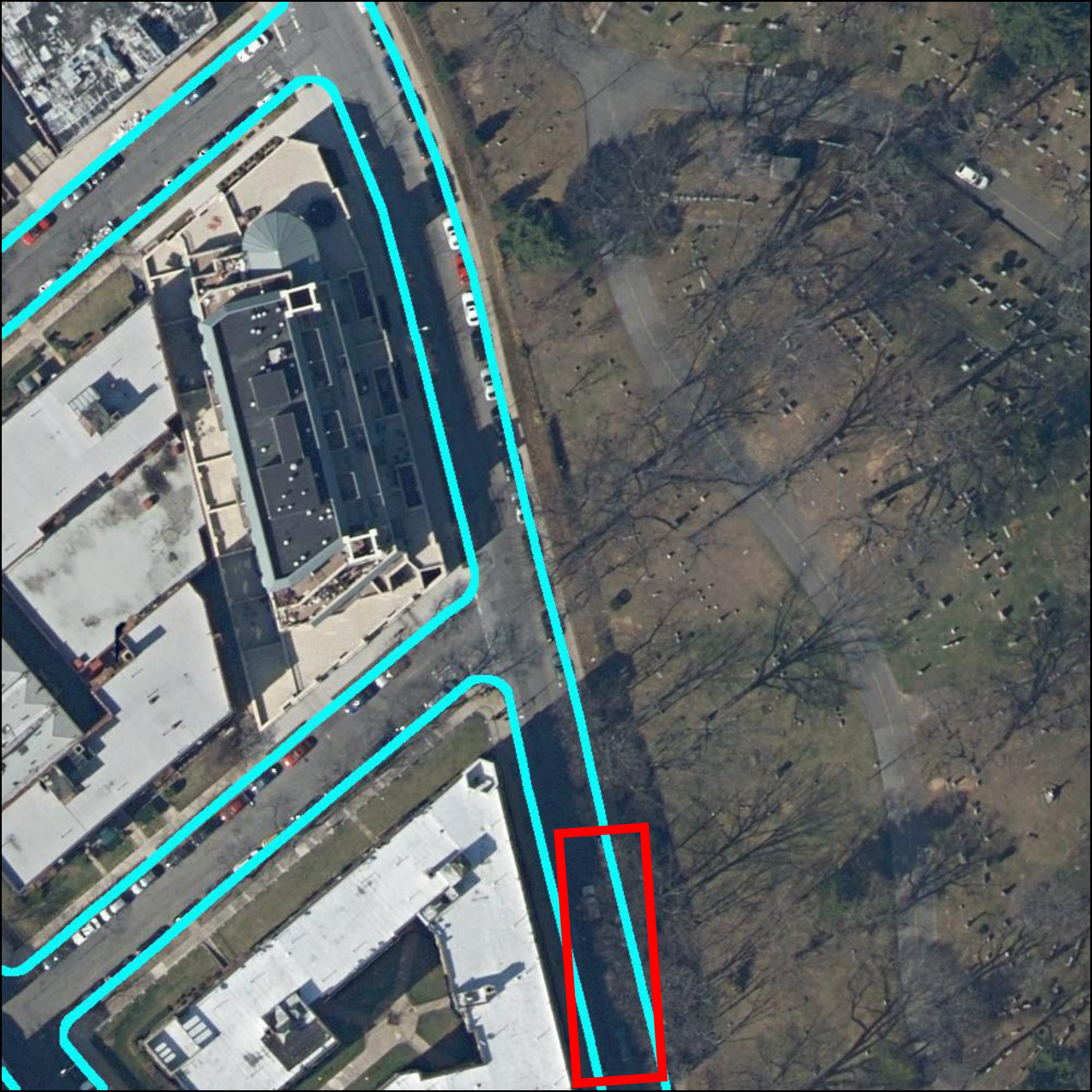}
        \end{subfigure}
        \caption{Ground-truth}
    \end{subfigure}
    \hfill
    \begin{subfigure}[t]{0.155\textwidth}
        \begin{subfigure}[t]{\textwidth}
            \includegraphics[width=\textwidth]{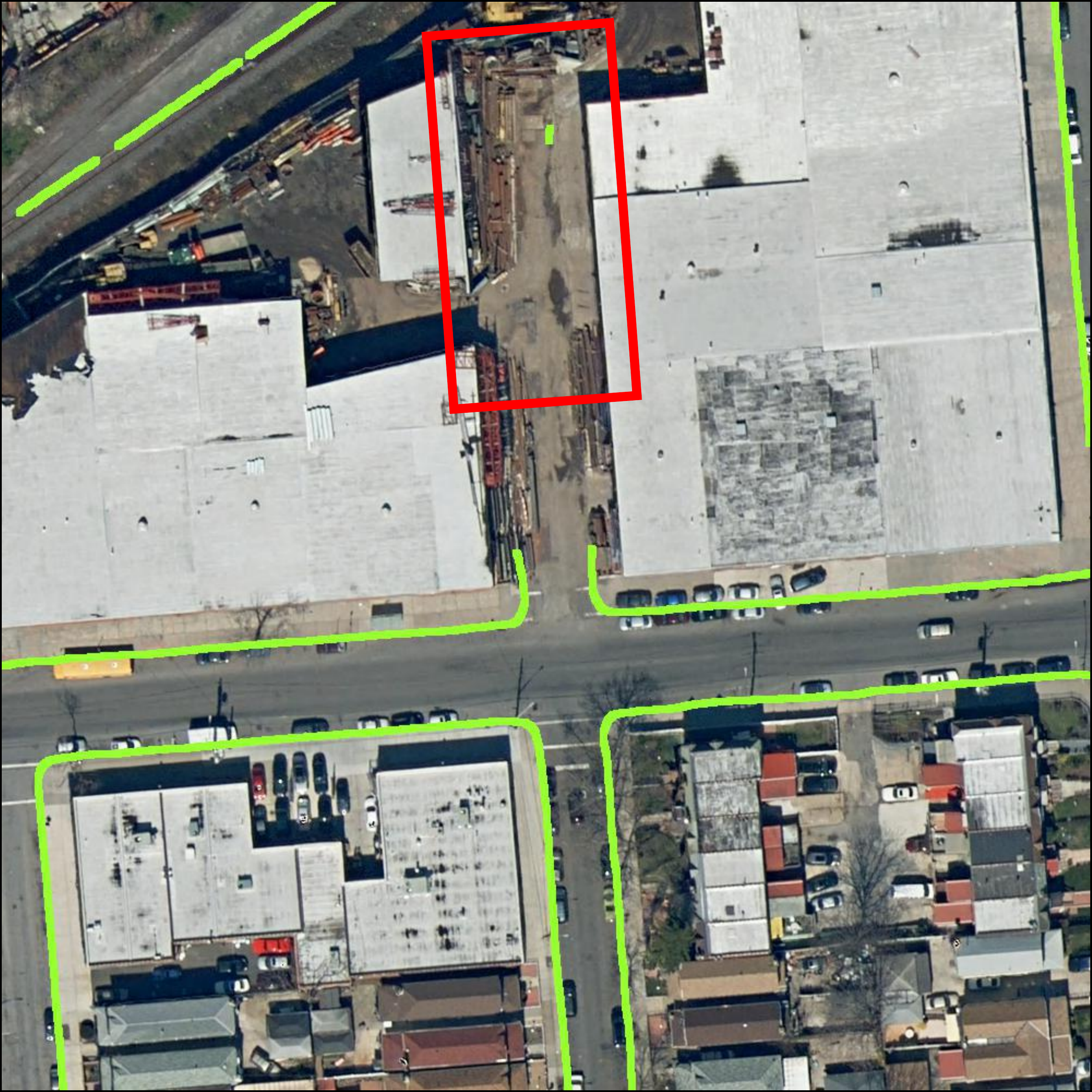}
        \end{subfigure}\vspace{.6ex}
        \begin{subfigure}[t]{\textwidth}
            \includegraphics[width=\textwidth]{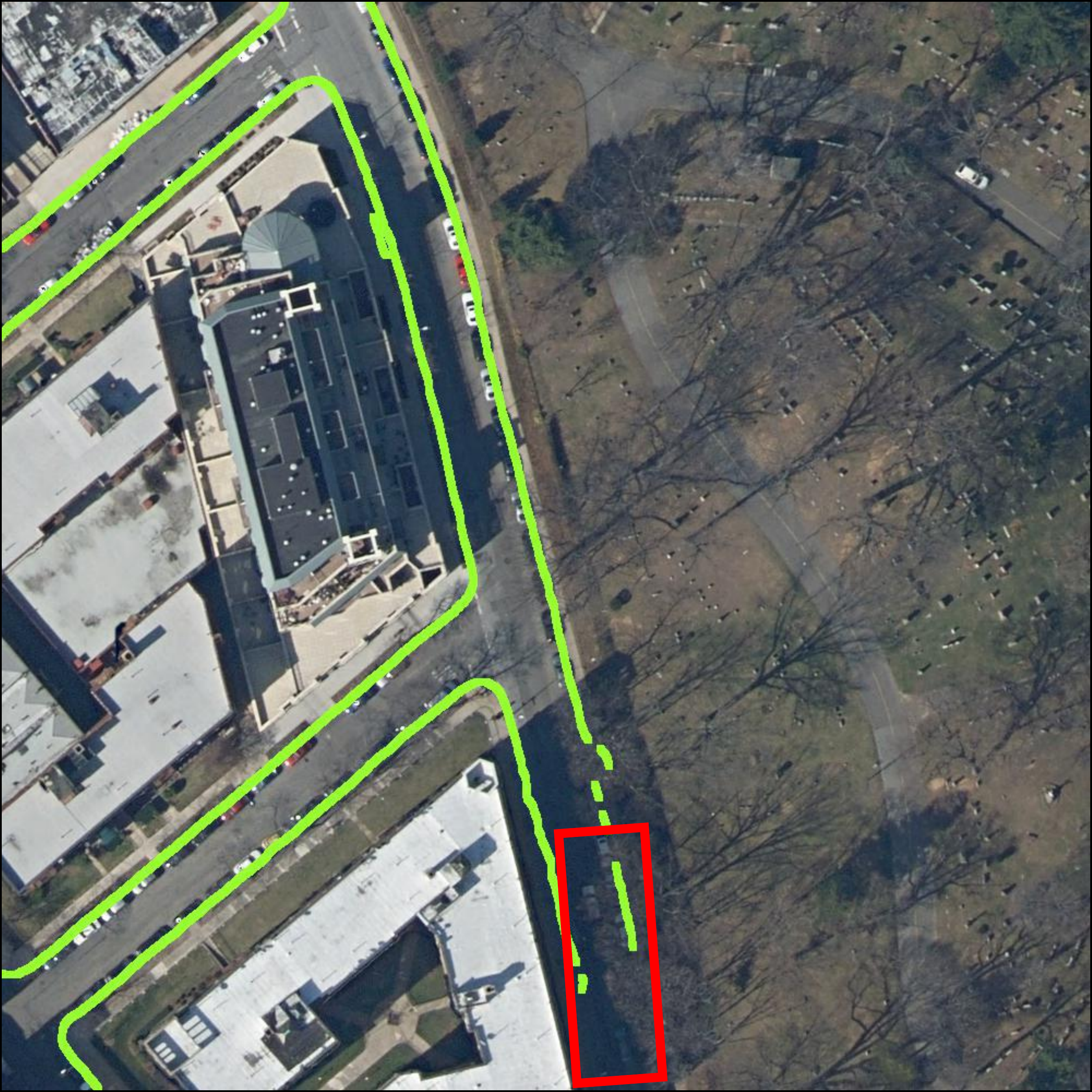}
        \end{subfigure}
        \caption{Naive baseline}
    \end{subfigure}
    \hfill
    \begin{subfigure}[t]{0.155\textwidth}
        \begin{subfigure}[t]{\textwidth}
            \includegraphics[width=\textwidth]{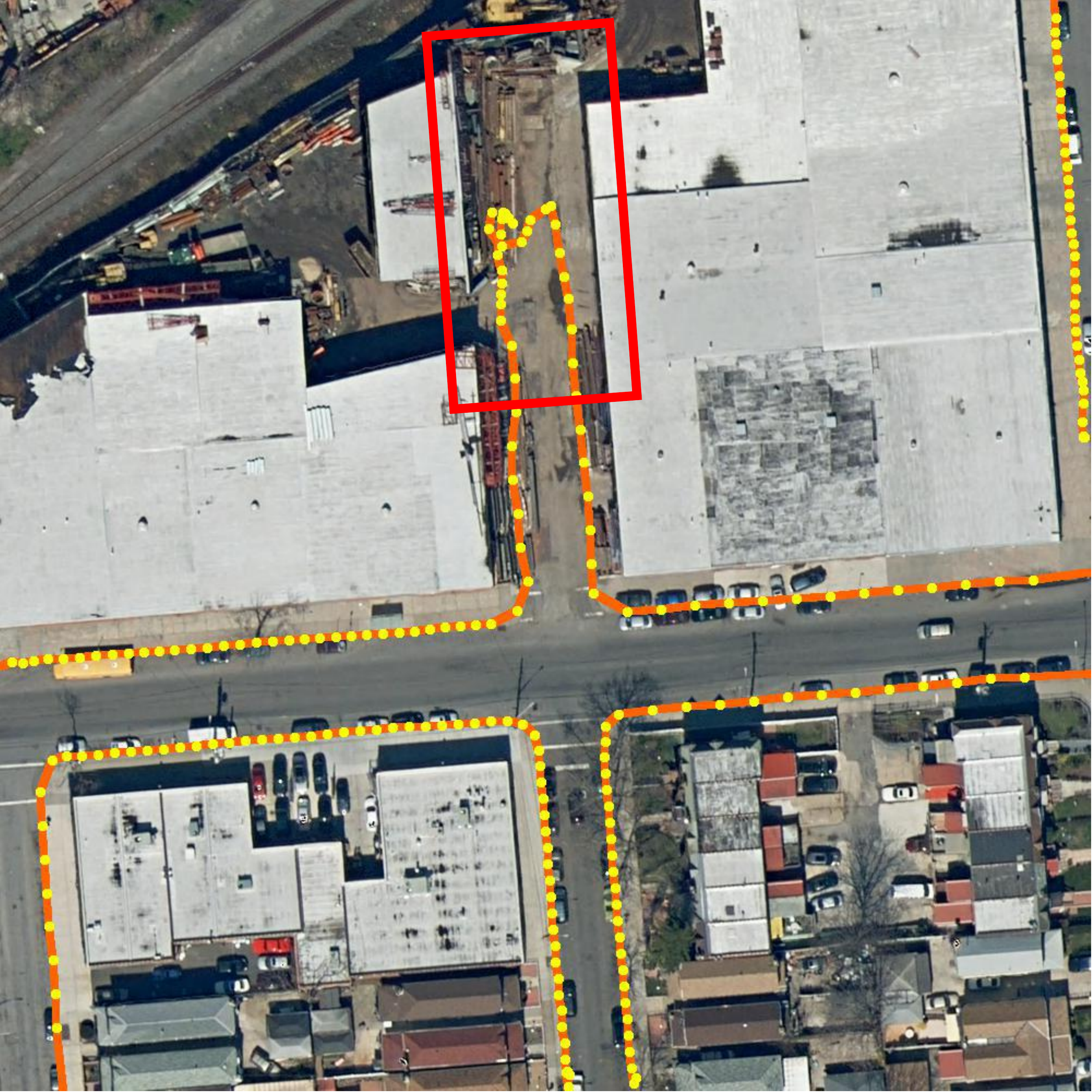}
        \end{subfigure}\vspace{.6ex}
        \begin{subfigure}[t]{\textwidth}
            \includegraphics[width=\textwidth]{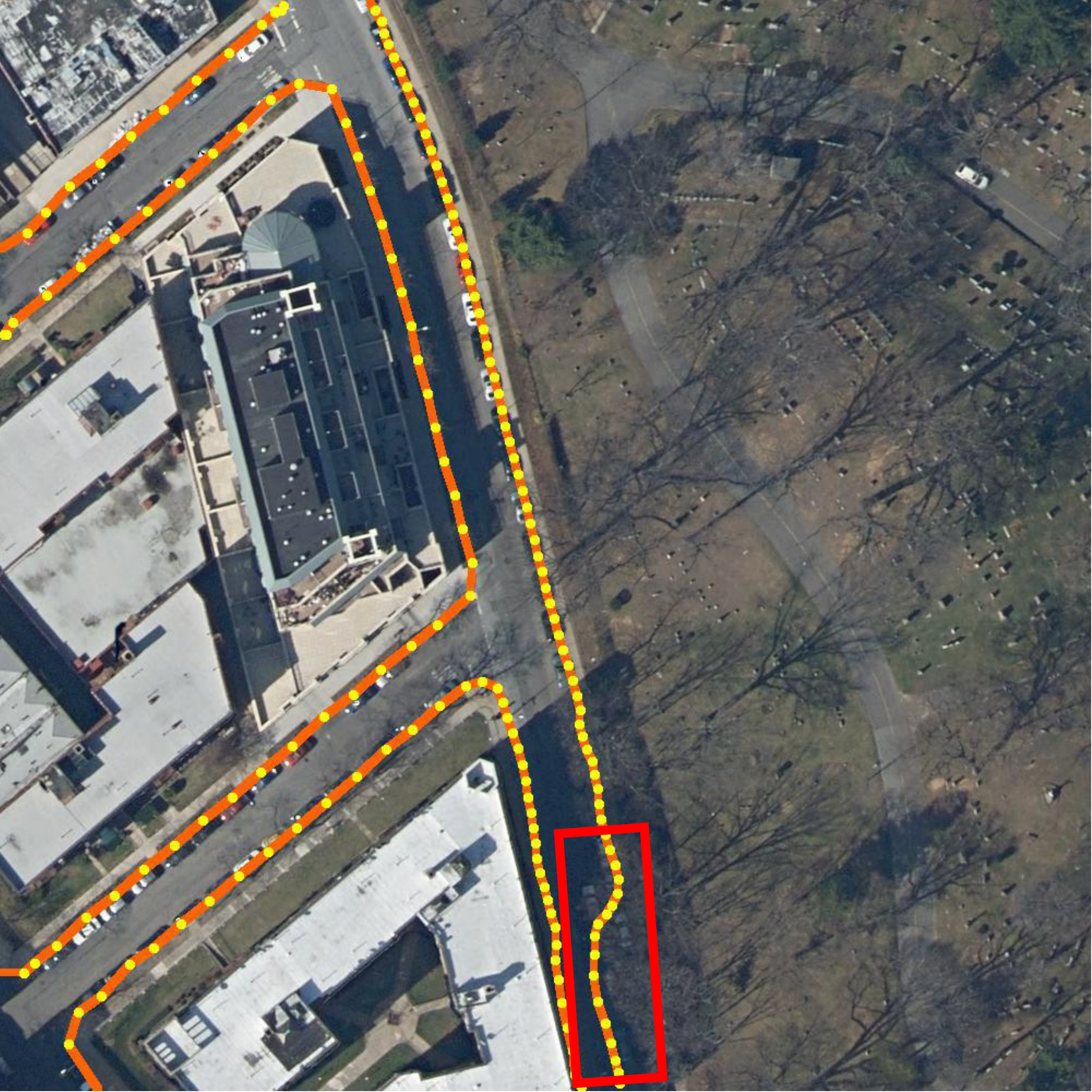}
        \end{subfigure}
        \caption{iCurb}
    \end{subfigure}
    \caption{Qualitative demonstrations for the failure cases. (a) The ground-truth (cyan lines); (b) The result of the naive baseline (green lines); (c) The result of iCurb (orange lines for edges and yellow nodes for vertices). Red rectangles indicate where the errors occur. The first row shows an example of unsatisfactory results on an unstructured road. Such a road is rare in the training set, so the agent $\bm{A}$ fails to generate correct graphs. The second row is an example of the unexpected agent behavior caused by tree occlusion. These errors could be relieved by using more powerful backbone networks in the future. The figure is best viewed in color.}
    \label{fig_failure_case}
\end{figure}

Under most cases, iCurb could handle various scenarios and produce satisfied road-curb graphs. From the experimental results, we notice that even if one head of $\bm{A}$ makes incorrect predictions, the other head could still guarantee $\bm{A}$ to output acceptable results. For example, if the coordinate head makes a wrong prediction of $x_{t+1}$ (e.g., far from the right track), the stop head would stop $\bm{A}$ immediately to prevent it from growing wrong curb graphs. Similarly, if the stop head does not predict a stop action correctly, the coordinate head would always predict $x_{t+1}$ the same as $x_t$ so that no incorrect vertices are added. However, when two heads fail at the same time, our iCurb would produce graphs with notable errors. Some examples are visualized in Fig. \ref{fig_failure_case}. 

\subsection{Limitations}
The limitations of iCurb are mainly three-fold. Firstly, the results of iCurb are relatively less smooth. This is because the vertices generated by iCurb are discrete so that connecting the vertices using lines makes the graph look less smooth. Secondly, the training process usually takes a long time. This is because iCurb will run $1+N$ rounds of explorations for every image, which is time-consuming.
Finally, iCurb fails to generate correct graphs when two heads of $\bm{A}$ have wrong predictions simultaneously. Such errors mainly occur when road curbs are severely occluded by shadows or plants. This issue might be relieved by utilizing more powerful backbone or multi-modal semantic segmentation networks \cite{sun2019rtfnet} in the future.

%%%%%%%%%%%% Figure 6  %%%%%%%%%%
%%%%%%%%%%%% Figure 6  %%%%%%%%%%

\section{Conclusions and Future Work}
We proposed here a novel solution for road-curb detection using aerial images for autonomous driving. We formulated our problem as an imitation learning problem and developed a novel network iCurb that takes as input aerial images and directly outputs road-curb graphs. 
In iCurb, we designed an agent to iteratively generate the graph. The proposed agent could make use of historical information and predict the next vertex. To enable the agent to explore larger state space and guarantee the quality of obtained graphs at the same time, we proposed an innovative training algorithm that consists of the restricted exploration method and the free exploration method. 
We created a new dataset based on the NYC-planimetrics dataset. The results demonstrated the effectiveness and superiority of our method, especially for the connectivity. 
In the future, we will try more powerful backbones or multi-modal semantic segmentation networks. We plan to explore better training strategies that are more efficient and effective. Besides, iCurb will be tested on new datasets for generalization-capability evaluation.

\bibliographystyle{IEEEtran}
\bibliography{mybib}
 
\end{document}